\documentclass[conference]{IEEEtran}
\IEEEoverridecommandlockouts
\usepackage{cite}
\usepackage{amsmath,amssymb,amsfonts}
\usepackage{algorithmic}
\usepackage{graphicx}
\usepackage{textcomp}
\usepackage{xcolor}
\usepackage{url}
\def\BibTeX{{\rm B\kern-.05em{\sc i\kern-.025em b}\kern-.08em
    T\kern-.1667em\lower.7ex\hbox{E}\kern-.125emX}}

\usepackage{tikz}
\usepackage{textcomp}
\usepackage{hyperref}
\usepackage{lipsum}

\newcommand\copyrighttext{%
  \footnotesize \textcopyright 2021 IEEE. Personal use of this material is permitted.
  Permission from IEEE must be obtained for all other uses, in any current or future
  media, including reprinting/republishing this material for advertising or promotional
  purposes, creating new collective works, for resale or redistribution to servers or
  lists, or reuse of any copyrighted component of this work in other works.\\
  \emph{Proceedings of the 2021 IEEE International Conference on Omni-layer Intelligent systems (IEEE COINS 2021)}}
\newcommand\copyrightnotice{%
\begin{tikzpicture}[remember picture,overlay]
\node[anchor=north,yshift=-5pt] at (current page.north) {\fbox{\parbox{\dimexpr\textwidth-\fboxsep-\fboxrule\relax}{\copyrighttext}}};
\end{tikzpicture}%
}

\makeatletter
\newcommand{\linebreakand}{%
  \end{@IEEEauthorhalign}
  \hfill\mbox{}\par
  \mbox{}\hfill\begin{@IEEEauthorhalign}
}
\makeatother

\begin{document}

\title{TEACHING - Trustworthy autonomous cyber-physical applications through human-centred intelligence
\thanks{This research was supported by TEACHING, a project funded by the EU Horizon 2020 research and innovation programme under GA n. 871385} 
}

\author{\IEEEauthorblockN{Davide Bacciu\IEEEauthorrefmark{1}, Siranush Akarmazyan\IEEEauthorrefmark{2}, Eric Armengaud\IEEEauthorrefmark{4}, Manlio Bacco\IEEEauthorrefmark{3}, George Bravos\IEEEauthorrefmark{2}, Calogero Calandra\IEEEauthorrefmark{8},\and Emanuele Carlini\IEEEauthorrefmark{3}, Antonio Carta\IEEEauthorrefmark{1}, Pietro Cassarà\IEEEauthorrefmark{3}, Massimo Coppola\IEEEauthorrefmark{3}, Charalampos Davalas\IEEEauthorrefmark{6}, Patrizio Dazzi\IEEEauthorrefmark{3}, \and Maria Carmela Degennaro\IEEEauthorrefmark{8}, Daniele Di Sarli\IEEEauthorrefmark{1}, Juergen Dobaj\IEEEauthorrefmark{5}, Claudio Gallicchio\IEEEauthorrefmark{1}, Sylvain Girbal\IEEEauthorrefmark{9},  \and Alberto Gotta\IEEEauthorrefmark{3}, Riccardo Groppo\IEEEauthorrefmark{10}, Vincenzo Lomonaco \IEEEauthorrefmark{1}, Georg Macher\IEEEauthorrefmark{5}, Daniele Mazzei\IEEEauthorrefmark{1}, Gabriele Mencagli\IEEEauthorrefmark{1}, \and Dimitrios Michail\IEEEauthorrefmark{6},  Alessio Micheli\IEEEauthorrefmark{1}, Roberta Peroglio\IEEEauthorrefmark{10}, Salvatore Petroni\IEEEauthorrefmark{8}, Rosaria Potenza\IEEEauthorrefmark{8}, Farank Pourdanesh\IEEEauthorrefmark{8}, \and Christos Sardianos\IEEEauthorrefmark{6},  Konstantinos Tserpes\IEEEauthorrefmark{6}, Fulvio Tagliab{\'o}\IEEEauthorrefmark{8}, Jakob Valtl\IEEEauthorrefmark{7}, Iraklis Varlamis\IEEEauthorrefmark{6} and Omar Veledar\IEEEauthorrefmark{4}} \linebreakand

\IEEEauthorblockA{\IEEEauthorrefmark{1}University of Pisa, Pisa, Italy, Email:\url{bacciu@di.unipi.it}}
\IEEEauthorblockA{\IEEEauthorrefmark{2}Information Technology for Market Leadership, Greece, Email: \url{gebravos@itml.gr}} \IEEEauthorblockA{\IEEEauthorrefmark{3}Institute of Information Science and Technologies (ISTI), CNR, Italy, Email: \url{patrizio.dazzi@isti.cnr.it}}
\IEEEauthorblockA{\IEEEauthorrefmark{4}AVL List GmbH, Graz, Austria, Email: \url{omar.veledar@avl.com}}
\IEEEauthorblockA{\IEEEauthorrefmark{5}Graz University of Technology, Graz, Austria, Email: \url{georg.macher@tugraz.at}}
\IEEEauthorblockA{\IEEEauthorrefmark{6}Harokopio University of Athens, Greece, Email: \url{tserpes@hua.gr}}
\IEEEauthorblockA{\IEEEauthorrefmark{7}Infineon Technologies AG, Munich, Germany, Email: \url{jakob.valtl@infineon.com}}
\IEEEauthorblockA{\IEEEauthorrefmark{8}Marelli Europe S.p.A, Turin, Italy, Email: \url{rosaria.potenza@marelli.com}}
\IEEEauthorblockA{\IEEEauthorrefmark{9}Thales Research and Technology, France, Email: \url{sylvain.girbal@thalesgroup.com}}
\IEEEauthorblockA{\IEEEauthorrefmark{10}Ideas \& Motion, Turin, Italy, Email: \url{riccardo.groppo@ideasandmotion.com}}
} 

\maketitle
\copyrightnotice

\begin{abstract}
This paper discusses the perspective of the H2020 TEACHING project on the next generation of autonomous applications running in a distributed and highly heterogeneous environment comprising both virtual and physical resources spanning the edge-cloud continuum. TEACHING puts forward a human-centred vision leveraging the physiological, emotional, and cognitive state of the users as a driver for the adaptation and optimization of the autonomous applications. It does so by building a distributed, embedded and federated learning system complemented by methods and tools to enforce its dependability, security and privacy preservation. The paper discusses the main concepts of the TEACHING approach and singles out the main AI-related research challenges associated with it. Further, we provide a discussion of the design choices for the TEACHING system to tackle the aforementioned challenges
\end{abstract}


\section{Introduction}
The world is on the verge of the autonomous systems revolution. Autonomous virtual agents handle customer care, bots autonomously process human discourse and generate targeted content for social networks communication, while autonomous vehicles are entering the industrial and commercial markets. Automation is the technology enabling the conduction of processes with minimum human assistance \cite{groover2010fundamentals}, which spells out as autonomy when the human is taken out of the sensing, decision and actuation loop. Automation can be used to operate complex systems comprising multi-faceted and dynamic virtual and physical resources, such as telecommunication networks, factories, ships, aircraft and vehicles, with minimal or reduced human intervention. Such systems living at the crossroads of the real and virtual world are referred to with the umbrella term of Cyber-physical Systems (CPS) \cite{cps}.

Even when the most advanced degree of autonomy is exercised, the human is an unavoidable variable for any safety-critical scenario. Humans interact with autonomous systems either as passive end-users of the service being delivered (such as passengers in autonomous vehicles) or as active co-operators in a mutual empowerment relationship towards a shared goal (e.g. in industrial assembly lines).  Such cooperative, connected, and autonomous systems of systems (SoS) are potential game-changers in multiple domains that are prepared to positively exploit such inescapable human factor.

Human-centric autonomous CPS exposes critical requirements in terms of adaptation capabilities and several degrees of trustworthiness (e.g. dependability and privacy). It also considers human comfort and distress throughout system operation. Nevertheless, it also enables unparalleled innovation potential throughout the realization of a holistic intelligent environment, where the human and the cyber-physical entities support, cooperate and, ultimately empower each other.

In other words, we are seeking solutions that heavily root into Artificial Intelligence (AI). AI is a key technology to realize autonomous applications, even more so when such applications are realized within the inherently dynamic, connected, and interacting context of a CPS \cite{Vermesan1162039}. Machine Learning (ML) models, in particular, allow dynamic acquisition of knowledge from historic data to anticipate the effect of actions, plans and interactions within the CPS, with the entangled physical environment and, ultimately, with the human. AI has the potential to become the key enabler of emerging cooperative behaviour between the human and the cyber-physical world. The stringent computational and memory requirements of AI impose a significant rethinking of the underlying computing software and system, which need to provide AI-specialized support in the computing fabric, even at a hardware level. Simultaneously, the realization of such intelligent empowerment of the CPS raises compelling challenges related to AI fundamentals, to the trustworthiness of AI-based systems and to their ability to cater and care for the human stakeholders.

The H2020 project TEACHING (grant n. 871385,2020-2022) is a recent 3-years research endeavour targeting specifically the provisioning of innovative methods and systems to enable the development of the next generation of autonomous AI-based applications distributed over CPSs. TEACHING puts forward a human-centric perspective on CPS intelligence based on a synergistic collaboration between human and cybernetic intelligence. The TEACHING concept is rooted in the long-standing concept of Humanistic Intelligence (HI)\cite{Minsky2013TheSO}. That is the intelligence that arises when a human being is placed in the feedback loop of a computational process. Within such a conceptual framework, TEACHING instantiates several AI methodologies comprising distributed learning systems, embedded recurrent neural models, federated learning, continual learning, learning under drifting distributions and from sparse feedback. Further, the TEACHING human-centric perspective brings up the issue of how to elicit the necessary feedback to drive adaptation in the right direction. When the human is in the loop, it is natural to consider him/her as a source of informative and dependable teaching information. However, explicit elicitation of human feedback has demonstrated to be an unsustainable way of achieving adaptation, typically causing the user to stop interacting with the system. In this respect, TEACHING also investigates novel, creative and reliable forms of human feedback collection and for their incorporation into learning models. 

In this paper, we outline the characterizing aspects of the TEACHING approach, which stem from providing an answer to the following research questions:
\begin{enumerate}
\item[Q1] How can we construct a cooperative human-CPS intelligent environment where the needs, comfort and well-being of the human are at the core of the CPS?
\item[Q2] How can such a cooperative environment be realized to operate in an autonomous and trustworthy way, while being capable of self-adapting by exploiting sustainable human feedback?
\item[Q3] How do we change the underlying computing system, at an architectural and software level, to support the operation of such an adaptive, dependable and human-centric CPS?
\end{enumerate}
Providing a compelling answer to these questions is fundamental for many safety-critical applications that are key in the European industrial landscape, such as automotive, avionics, and general autonomous transportation, which are the main project use cases.
In the following, we discuss the fundamental concepts underlying the TEACHING response to these fundamental questions (Section \ref{sect:approach}). We then outline the conceptual architecture of the TEACHING solution (Section \ref{sect:arch}) and conclude with a discussion of the main AI-related challenges of the TEACHING vision and the methodology put forward by the project to tackle those challenges (Section \ref{sect:disc}).

Due to space limitations, this paper focuses mainly on the TEACHING project concepts and contributions related to distributed learning on the cloud-edge continuum and AI-as-a-service (AIaaS) for autonomous CPS applications, providing only a synthetic view over the other relevant components of the TEACHING system.

\section{The TEACHING approach} \label{sect:approach}
TEACHING develops a human-centric CPS for autonomous safety-critical applications based on a distributed, energy-efficient and trustworthy AI, leveraging specialized computing fabric for AI and in-silico support for intelligent cybersecurity solutions. AI methodologies will be developed to support the design and deployment of autonomous, adaptive and dependable CPS applications, allowing the exploitation of sustainable human feedback to drive, optimize and personalize services.  TEACHING devises an intelligent environment where the human and the cybernetic entities collaborate synergistically, where the latter provides the former with a comfortable, tailored and dependable interaction driven by the implicit feedback provided by the human throughout his/her physiological reactions to CPS operation.

\subsection{Cooperative human-centric autonomy (Q1).} TEACHING advances a human-aware approach, where human reactions are monitored, in an unobtrusive, respectful, and privacy-preserving fashion, to evaluate and inform the operation of the autonomous CPS. Human reactions are monitored in the least obtrusive way, without disrupting human attention from the primary operation they are performing. TEACHING comprises a specialized human-centric system of sensing devices integrated within the CPS and encompassing wearable and environmental sensors. These serve as information sources feeding AI models specialized in the recognition and characterization of the human physiological, emotional, and cognitive (PEC) state \cite{colombo,nca2020}. The reactions monitored by such a system will drive CPS operation in synergy with the humans.

\subsection{Autonomous trustworthy adaptation by HI (Q2).}
TEACHING builds on AI as a key enabler for autonomous CPS with integrated decision-making capabilities. TEACHING will provide a toolkit implementing AI-as-a-service (AIaaS) functionalities \cite{iotML} for the CPS, allowing components and applications to access data-driven intelligence, adaptivity and prediction mechanisms, seamlessly. TEACHING embraces a view targeting an AI rooted in the HI concepts, considering human feedback as a crucial driver to evaluate the operation of the CPS AI but also to allow its adaptation. To this end, TEACHING will develop the necessary methodology to allow the CPS AI to self-adapt and personalize, exploiting the human physiological, emotive or cognitive (PEC) state as a non-obtrusive source of teaching information. The connected and autonomous nature of TEACHING poses challenging demands of dependability and security, reinforced by requirements induced by its signature human-aware approach and by the widespread use of AI technologies (e.g. privacy). TEACHING explicitly addresses such crucial aspects through a solution that acts synergistically on the nature of AI models and of the computing system. First, we take an edge-distributed and federated AI approach, maintaining important parts of the computation close to the end-user and the data source, reducing connectivity-related threats to reliability and security, while enabling the exploitation of virtually endless cloud resources. Second, TEACHING explicitly addresses the risks of hampered or deteriorating AI models subject to attacks \cite{BIGGIO2018317} or continuous learning drifts that \cite{Wcci20CL}, providing mechanisms for early detection and replacement with certified AI models. Privacy-aware AI methodologies \cite{privacy} will be bundled within the AIaaS to avoid exposing sensitive and critical information.

\subsection{Supporting trustworthy AI at the computing level (Q3).}
TEACHING develops a computing system and middleware, whose design is guided by the human-aware, AI-related, and application-driven needs discussed above. It develops a high-performance edge and federated computing infrastructure able to efficiently support the demanding AI models that will deliver the desired HI in the CPS but also offering native in-silico support to dependability and cyber-security requirements. To this end, TEACHING devises a decentralized computing approach comprising heterogeneous technologies, including solutions based on specialized hardware (e.g. in-silico AI).
TEACHING middleware complements the high heterogeneity and specificity of the hardware resources with an abstraction, communication and orchestration layer employing approaches from the cloud- and edge-computing, enabling the management of resources and applications onto a computing continuum spanning the whole platform. TEACHING will deliver proper abstractions that fit with the programming issues and needs of AI/ML models, whose implementation should be provided on a variety of different resources (silicon-AI, multicores, GPUs and FPGAs). Similar abstractions will enable the specification and implementation of trustworthiness-related aspects.

\section{TEACHING architecture} \label{sect:arch}

\subsection{Conceptual design}
\begin{figure}[tb]
    \centering
    \includegraphics[width=.40\textwidth]{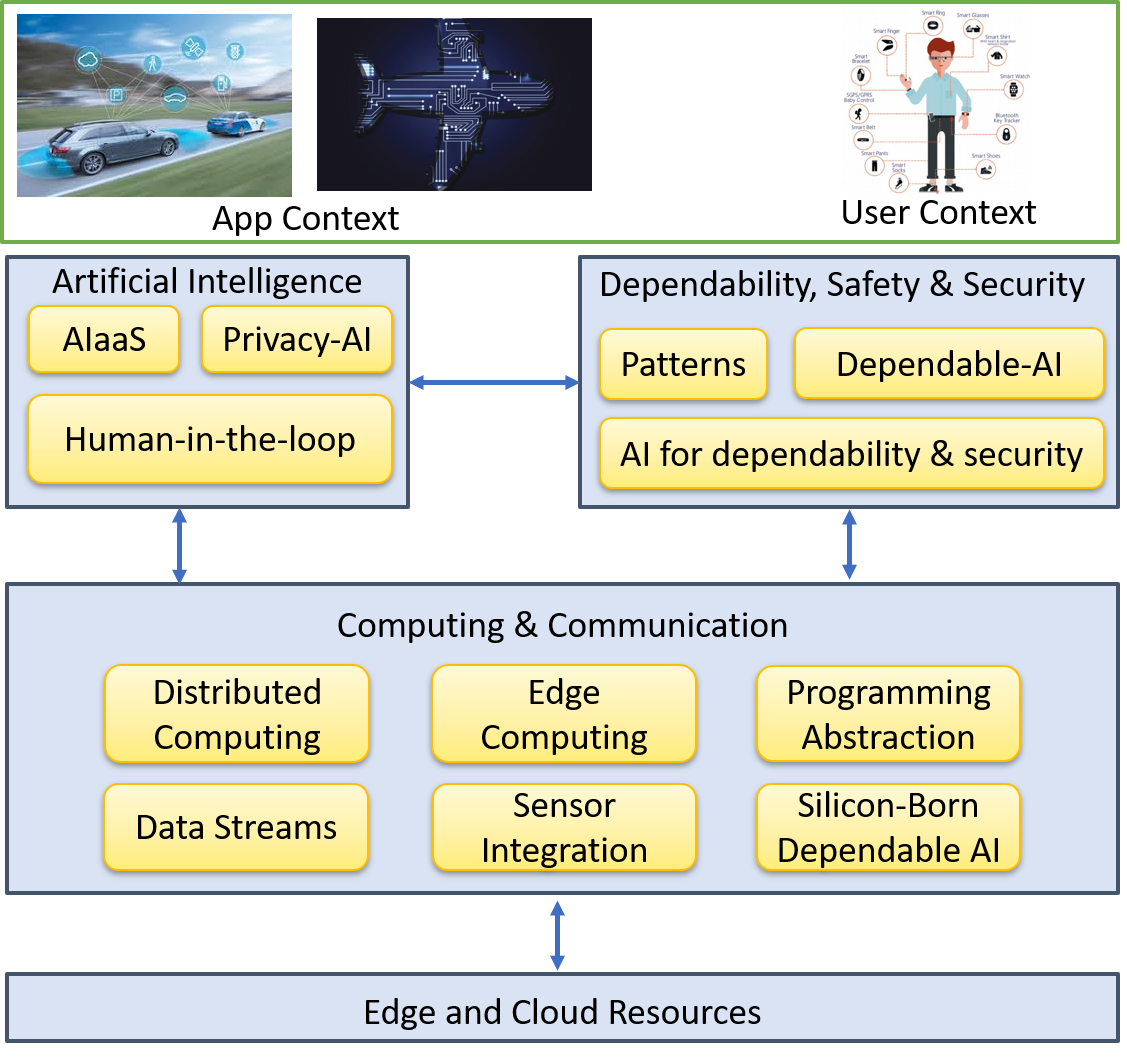}
    \caption{TEACHING abstract architecture with its three main subsystems.}
    \label{fig:arch}
\end{figure}

The TEACHING solution builds on a conceptual design, which is exemplified by the abstract architecture in Figure \ref{fig:arch}, composed of three main subsystems. At the bottom is {\it Computing \& Communication}, a mixed edge/cloud orchestration platform, abstracting from the heterogeneous nature of the underlying resources. The virtualized nodes are expected to bear variable privacy, availability, reliability, security, and latency properties. To this end, the platform caters for the optimized management of specialized resources that are of interest to the AI applications and the management of the prospective high-frequency data streams. It will also lay the foundations for the deployment and migration of AI tasks. {\it Dependability, safety and security} consider trustworthiness across all engineering phases and at runtime. The platform provides a collection of engineering methods, design patterns, and mechanisms, which ensure that the CPS will provide end-to-end trustworthiness guarantees to the AI-based applications running in the CPS. The {\it Artificial Intelligence} subsystem includes several underlying components capable of collecting non-obtrusive feedback from the human that consumes the AI-based service. It leverages embedded and edge computing AI solutions and provides mechanisms to support AI autonomous applications in an AIaaS fashion. These mechanisms also enable personalizing the AI models to the particular human beings monitored and ensure that sensitive information will remain private.

\subsection{The TEACHING platform}

\begin{figure}[tb]
    \centering
    \includegraphics[width=.5\textwidth]{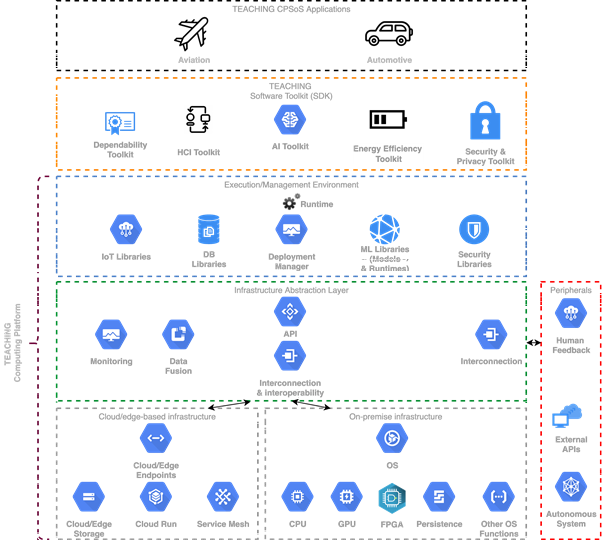}
    \caption{High-level design of the TEACHING Platform.}
    \label{fig:plat}
\end{figure}

The TEACHING Platform is the combined stack of the computing platform and software toolkit upon which a developer develops and deploys CPSoS applications. A high-level view of this platform is provided by Figure \ref{fig:plat}, following the rationale of the layered conceptual design where each layer offers services to the one above. An instance of this architecture may include implementations that merge layers, similarly as ISO/OSI and TCP/IP.

The TEACHING platform is comprised of 5 layers, each of which provides services to the one above. At the bottom of the stack, we have the {\it Infrastructure Layer} which is comprised of various heterogeneous infrastructures, exposed through an embedded system OS and the cloud/edge resources. TEACHING assumes that access to the resources of those infrastructures is a priori possible. On that premise, the first task of TEACHING is to homogenize those resources, something that is the main functionality of the {\it Infrastructure Abstraction Layer} (IAL). The IAL homogenizes the underlying infrastructures providing a single API to deploy, execute and monitor resources and application components. This layer also caters for implementing I/Os, with the underlying persistence layers as well as with the supported peripherals, i.e., the target autonomous system (CPS) and external APIs (e.g., web services). The {\it Execution/Management Environment} (EME) exposes a single API that facilitates the execution and lifecycle management of the application components. It provides the runtime for that purpose, along with integrated libraries, implemented at a low-abstraction language, providing services and optimizations at the top layers. Such libraries include ML runtimes such as those of Tensorflow and PyTorch, or ML optimizations in Python and C++. It also includes libraries for managing IoT solutions (e.g., OS-IoT) implementing IoT protocols such as OneM2M. Other libraries include the DB and security libraries, ensuring that such functionality is provided to the layers above.

The {\it TEACHING Software Toolkit} (SDK) provides the framework and APIs to implement CPSoS applications making the best use of the CPSoS services. The TEACHING SDK supports 6 toolkits:
\begin{itemize}
    \item The AI toolkit is the software library that allows the developer to invoke learning modules, set up training or inference procedures, etc. The AI toolkit has the appropriate wirings with the underlying layers to deploy and run the ML components at the appropriate resources (e.g., GPUs) and facilitates the I/Os and dataset management.
    \item The HCI toolkit allows the software developer to invoke the services that are relevant to the human feedback, e.g., filters, buffers and other suchlike tools for retrieving and managing the human feedback. Furthermore, this toolkit includes design patterns and guidelines for human-centred design.
    \item The Security and Privacy toolkit provides readily available security APIs as well as privacy guidelines. In terms of security, the developers may define a part of their code or a standalone component that has to run on a secure enclave or that the communication between components has to use OpenSSL calls. In terms of privacy, the developers may identify datasets as containing sensitive data, thus implicitly imposing constraints in their further use. Furthermore, the privacy toolkit may also include functional tools like anonymizers.
    \item The Dependability toolkit provides software that audits the code or application components against the TEACHING dependability guidelines/procedures. It also provides engineering patterns implementations that the developers can invoke for ensuring the dependable execution of software. For instance, in cases where the developers invoke online training approaches through the AI toolkit, the dependability toolkit may allow the code to run in multiple instances implementing a consensus model.
    \item The Energy Efficiency (EE) toolkit is linking the code or components that the user would like to run with EE services provided by the underlying layers. E.g., in order to run an application, the toolkit may employ energy-efficient approaches such as dynamic voltage and frequency scaling (DVFS), power mode management (PMM) or unconventional cores such as DSP or GPUs of FPGAs. This can be done automatically or invoked by the user (e.g., by using code “annotations”).
\end{itemize}

The final layer of this architecture relates to the {\it TEACHING CPSoS Applications}, which may be comprised of loosely coupled, standalone, independent components (e.g., docker images) that the TEACHING SDK builds or software that the TEACHING SDK compiles and executes.


\section{TEACHING AI challenges and methodology} \label{sect:disc}
Developing human-centred intelligence in CPS applications poses fundamental AI-related challenges crucial to the TEACHING vision. In the following, we provide a summary of the main challenges tackled by TEACHING along with a brief introduction of the methodologies being developed to address them.

\subsection{Fast learning in temporal domains}
The first key decision to be taken in designing a distributed learning system is the nature of the learning machinery to be used. The solution put forward by TEACHING has been determined based on considerations related to (i) the nature of the data that we expect the system to process and (ii) the nature of the underlying computing system. As pertains to the former aspect, we recognize that TEACHING data have primarily a dynamic, time-evolving nature, consistent with the definition of time-series data. This is compound information ${\bf u} = {\bf u}(1),\dots,{\bf u}(t),\dots, {\bf u}(T)$ comprising observations ${\bf u}(t)$ captured at different times $t$, such as data captured by environmental/wearable sensors or event-based information flowing in an ICT system.  The family of learning models that appears more apt to process time-series of heterogeneous nature is that of the Recurrent Neural Networks (RNNs), which provide a neural model enriched by a dynamic neural memory that can store and compress the history of the observations supplied in input to the model.

As regards the second aspect, one needs to carefully consider the distributed nature of the CPS hosting the learning system. In this context, the efficiency of the learning model becomes a key requirement, in particular when considering an edge-distributed learning system where the challenge posed by the low-power, low-memory, battery-powered devices adds to the equation. Within the scope of the RNN models, the Reservoir Computing (RC) paradigm \cite{LUKOSEVICIUS2009127} allows for achieving exceptionally fast training times, which come with high energy efficiency and contained memory requirements.

\begin{figure}
\centering
    \includegraphics[width=.5\textwidth]{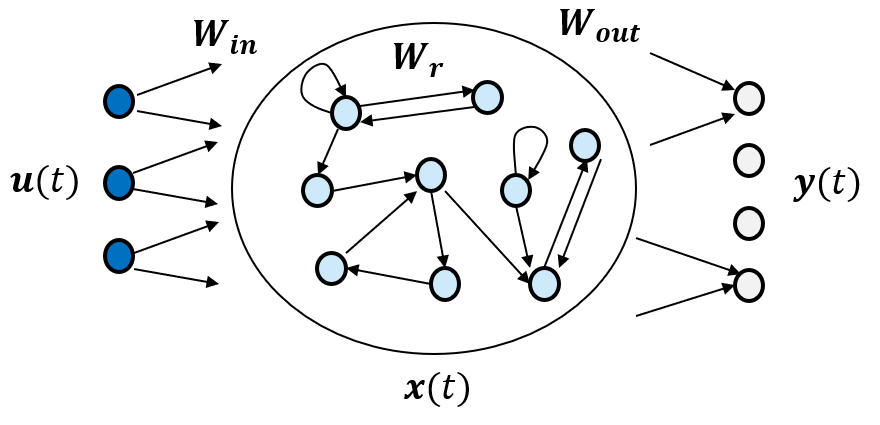}
    \caption{Architecture of an ESN: the input ${\bf u}(t)$ at time $y$ is fed to the recurrent reservoir, a state ${\bf x}(t)$ is computed and leveraged to produce the output ${\bf y}(t)$.}
    \label{fig:esn}
\end{figure}

RC provides a state-of-the-art trade-off between efficiency and effectiveness, \cite{BacciuBCGM14} which is due to the fact that the parameters of the recurrent part of the models are not learned. Instead, the properties of the underlying dynamical system are exploited to provide meaningful dynamic memory of the time-series even without training. The most popular instance of the RC paradigm is the Echo State Network (ESN) \cite{Jaeger78}, whose general architecture is described in Figure \ref{fig:esn}. It comprises a recurrent layer, the \emph{reservoir}, which holds an internal state ${\bf x}(t)$ that evolves over the time steps, and a \emph{readout}, which is a linear layer that transforms the reservoir state into a prediction ${\bf y}(t)$. The whole ESN is parameterized by $3$ matrices: ${\bf W}_{in}$ is the input-to-reservoir weight matrix, ${\bf W}_r$ is the recurrent reservoir-to-reservoir weight matrix, and ${\bf W}_{out}$ is the readout matrix. Only the readout matrix is trained (typically with a closed-form solution to a linear system) whereas the former two are randomly initialized and left untrained.

Motivated by such an efficiency-efficacy tradeoff, we have selected ESNs as the basic building blocks for the TEACHING learning systems. In particular,ESNs have been shown to scale from physical implementations \cite{TANAKA2019100}, to embedded applications on tiny devices \cite{BacciuBCGM14}, up to more powerful computing devices (e.g. cloud-based) in their deep reservoir version \cite{GALLICCHIO201787}. This choice is not only motivated by computational considerations. As it will become clearer in the next sections, ESNs are also characterized by appealing properties when it comes to designing learning mechanisms for distributed computing, such as with federated and continual learning.

\subsection{Federated learning}
Federated learning \cite{fedlearn} is essential to amalgamate the sharing concept with collaborative and distributed learning. Scattered multiple ML deployments generate localized individual model updates, which need to be kept coherent between each other and with a global model while accounting for the protection of personal and critical data of the human stakeholder.  The typical Federated Learning scenario comprises edge-based learning models trained on data available locally to the edge device that is periodically transferred to a cloud-resource where they are aggregated into a global model, typically by some form of weight averaging.  The choice of the aggregation strategy is critical and typically devoid of any guarantee about the quality of the aggregate model, especially when it comes to RNNs.

The TEACHING project is investigating the development of federated learning mechanisms designed specifically for ESN models. In particular, some early results of the project \cite{BacciuIJCNN2021} are showing how the use of ESNs enables a federation with guarantees of optimality. In particular, it is possible to devise an aggregation mechanism such that the fusion of the edge models in the federation is equivalent to training a centralized model using all the data available locally to the edge devices. By this means it is possible to realize a federated learning deployment with an excellent trade-off between accuracy and privacy preservation (as data does not need to be communicated out of the edge device where it is produced).

The TEACHING project will also be exploring the integration of zero-shot-learning methods \cite{zeroshot} and deep learning for graphs \cite{dgn20} to provide a unified framework to solve cold-start problems by integrating adaptive methods with prior knowledge (e.g. network topology, knowledge graphs, etc.).

\subsection{Continual learning on streams.} Traditional offline learning methods cannot smoothly adapt to new environmental conditions and drifting task distributions \cite{nonstatEnv}. Continual Learning (CL) focuses on the design of new models and techniques able to learn new information while preserving existing knowledge. Successfully tackling continual adaptation in the sequential/temporal domain (while retaining knowledge learned previously) is a key to success in autonomous CPS applications. Unfortunately, most of the consolidated continual learning strategies in literature seem poorly effective when applied to fully-adaptive RNNs \cite{cl4RNN}. Again, the TEACHING design choice of relying on untrained recurrent models has the potential to allow a neater application of CL strategies to the sequential domain. Preliminary project results \cite{cl4ESN} provide an evaluation of catastrophic forgetting in ESNs highlighting how they allow to successfully adapt a CL strategy for convolutional networks. Such strategy relies on the availability of a fixed feature extractor, that is the ESN untrained reservoir and that does not have a suitable counterpart in fully adaptive RNNs.

\subsection{Trustworthy distributed AI.}
The tight coupling between the autonomous application and the human poses high challenges on building trustworthy AI. We put forward a vision founding on using dependability engineering methods and design patterns for guaranteeing safety and dependability requirements in AI-based systems \cite{D3.1, Safecomp21}. We enhance the approach with runtime safeguards using continual monitoring of the human PEC state.

\subsection{Leveraging human state monitoring.}
The human PEC state during the interaction with an intelligent machine provides precious feedback on the machine's performance, which can be used for improving the intelligent machine itself. For such activity, the most promising psychological state to monitor is the distress level. Physiological signals (e.g. heart rate or skin conductivity) act as a proxy for the distress state \cite{CHEN2017279}, which machine learning algorithms can be trained to recognize. Trade-offs must be taken concerning the collection of the signals: while more signal usually provide a higher level of accuracy, invasiveness must be limited. Also, understanding the PEC state of a subject is a highly challenging task, as states can be subjective, difficult to characterize by the subject themselves, but ESNs have been shown to be effective in tasks involving the processing of physio-signals for human state monitoring \cite{colombo,rnn4HSM}.  Leveraging PEC state as a source of training feedback calls for novel methodologies that personalise autonomous driving using a combination of driving profiles and reinforcement learning techniques. This allows to optimise the vehicle behaviour and keep driver distress at a low level while driving within safety limits.



\begin{thebibliography}{10}
\providecommand{\url}[1]{#1}
\csname url@samestyle\endcsname
\providecommand{\newblock}{\relax}
\providecommand{\bibinfo}[2]{#2}
\providecommand{\BIBentrySTDinterwordspacing}{\spaceskip=0pt\relax}
\providecommand{\BIBentryALTinterwordstretchfactor}{4}
\providecommand{\BIBentryALTinterwordspacing}{\spaceskip=\fontdimen2\font plus
\BIBentryALTinterwordstretchfactor\fontdimen3\font minus
  \fontdimen4\font\relax}
\providecommand{\BIBforeignlanguage}[2]{{%
\expandafter\ifx\csname l@#1\endcsname\relax
\typeout{** WARNING: IEEEtran.bst: No hyphenation pattern has been}%
\typeout{** loaded for the language `#1'. Using the pattern for}%
\typeout{** the default language instead.}%
\else
\language=\csname l@#1\endcsname
\fi
#2}}
\providecommand{\BIBdecl}{\relax}
\BIBdecl

\bibitem{groover2010fundamentals}
M.~Groover, \emph{Fundamentals of Modern Manufacturing: Materials, Processes,
  and Systems}.\hskip 1em plus 0.5em minus 0.4em\relax John Wiley \& Sons,
  2010.

\bibitem{cps}
R.~{Rajkumar}, I.~{Lee}, L.~{Sha}, and J.~{Stankovic}, ``Cyber-physical
  systems: The next computing revolution,'' in \emph{Design Automation
  Conference}, 2010, pp. 731--736.

\bibitem{Vermesan1162039}
O.~Vermesan, A.~Br{\"o}ring, E.~Tragos, M.~Serrano, D.~Bacciu, S.~Chessa,
  C.~Gallicchio, A.~Micheli, M.~Dragone, A.~Saffiotti, P.~Simoens, F.~Cavallo,
  and R.~Bahr, ``Internet of robotic things : converging sensing/actuating,
  hypoconnectivity, artificial intelligence and iot platforms,'' in
  \emph{Cognitive Hyperconnected Digital Transformation : Internet of Things
  Intelligence Evolution}, 2017, pp. 97--155.

\bibitem{Minsky2013TheSO}
M.~Minsky, R.~Kurzweil, and S.~Mann, ``The society of intelligent veillance,''
  \emph{2013 IEEE International Symposium on Technology and Society (ISTAS):
  Social Implications of Wearable Computing and Augmediated Reality in Everyday
  Life}, pp. 13--17, 2013.

\bibitem{colombo}
D.~Bacciu, M.~Colombo, D.~Morelli, and D.~Plans, ``Randomized neural networks
  for preference learning with physiological data,'' \emph{Neurocomputing},
  vol. 298, pp. 9 -- 20, 2018.

\bibitem{nca2020}
D.~Bacciu, G.~Bertoncini, and D.~Morelli, ``Randomized neural networks for
  preference learning with physiological data,'' \emph{Neural Computing
  Applications}, 2021.

\bibitem{iotML}
D.~Bacciu, S.~Chessa, C.~Gallicchio, and A.~Micheli, ``On the need of machine
  learning as a service for the internet of things,'' in \emph{Proc. of the 1st
  Int. Conf. on Internet of Things and Machine Learning}, ser. IML '17.\hskip
  1em plus 0.5em minus 0.4em\relax ACM, 2017.

\bibitem{BIGGIO2018317}
B.~Biggio and F.~Roli, ``Wild patterns: Ten years after the rise of adversarial
  machine learning,'' \emph{Pattern Recognition}, vol.~84, pp. 317 -- 331,
  2018.

\bibitem{Wcci20CL}
A.~Cossu, A.~Carta, and D.~Bacciu, ``Continual learning with gated incremental
  memories for sequential data processing,'' in \emph{Proceedings of the 2020
  IEEE World Congress on Computational Intelligence}, 2020.

\bibitem{privacy}
N.~{Papernot}, P.~{McDaniel}, A.~{Sinha}, and M.~P. {Wellman}, ``Sok: Security
  and privacy in machine learning,'' in \emph{2018 IEEE European Symposium on
  Security and Privacy (EuroS P)}, 2018, pp. 399--414.

\bibitem{LUKOSEVICIUS2009127}
M.~Lukoševičius and H.~Jaeger, ``Reservoir computing approaches to recurrent
  neural network training,'' \emph{Computer Science Review}, vol.~3, no.~3, pp.
  127 -- 149, 2009.

\bibitem{BacciuBCGM14}
D.~Bacciu, P.~Barsocchi, S.~Chessa, C.~Gallicchio, and A.~Micheli, ``An
  experimental characterization of reservoir computing in ambient assisted
  living applications,'' \emph{Neural Computing and Applications}, vol.~24,
  no.~6, pp. 1451--1464, 2014.

\bibitem{Jaeger78}
H.~Jaeger and H.~Haas, ``Harnessing nonlinearity: Predicting chaotic systems
  and saving energy in wireless communication,'' \emph{Science}, vol. 304, no.
  5667, pp. 78--80, 2004.

\bibitem{TANAKA2019100}
G.~Tanaka, T.~Yamane, J.~B. Héroux, R.~Nakane, N.~Kanazawa, S.~Takeda,
  H.~Numata, D.~Nakano, and A.~Hirose, ``Recent advances in physical reservoir
  computing: A review,'' \emph{Neural Networks}, vol. 115, pp. 100--123, 2019.

\bibitem{GALLICCHIO201787}
C.~Gallicchio, A.~Micheli, and L.~Pedrelli, ``Deep reservoir computing: A
  critical experimental analysis,'' \emph{Neurocomputing}, vol. 268, pp.
  87--99, 2017.

\bibitem{fedlearn}
Q.~Yang, Y.~Liu, T.~Chen, and Y.~Tong, ``Federated machine learning: Concept
  and applications,'' \emph{ACM Trans. Intell. Syst. Technol.}, vol.~10, no.~2,
  Jan. 2019.

\bibitem{BacciuIJCNN2021}
D.~Bacciu, D.~D. Sarli, P.~Faraji, C.~Gallicchio, and A.~Micheli, ``Federated
  reservoir computing neural networks,'' in \emph{Proceedings of the
  International Joint Conference on Neural Networks (IJCNN 2021)}.\hskip 1em
  plus 0.5em minus 0.4em\relax IEEE, 2021.

\bibitem{zeroshot}
R.~Socher, M.~Ganjoo, C.~D. Manning, and A.~Ng, ``Zero-shot learning through
  cross-modal transfer,'' in \emph{Advances in Neural Information Processing
  Systems}, C.~J.~C. Burges, L.~Bottou, M.~Welling, Z.~Ghahramani, and K.~Q.
  Weinberger, Eds., vol.~26.\hskip 1em plus 0.5em minus 0.4em\relax Curran
  Associates, Inc., 2013, pp. 935--943.

\bibitem{dgn20}
D.~Bacciu, F.~Errica, A.~Micheli, and M.~Podda, ``A gentle introduction to deep
  learning for graphs,'' \emph{Neural Networks}, vol. 129, pp. 203 -- 221,
  2020.

\bibitem{nonstatEnv}
G.~{Ditzler}, M.~{Roveri}, C.~{Alippi}, and R.~{Polikar}, ``Learning in
  nonstationary environments: A survey,'' \emph{IEEE Computational Intelligence
  Magazine}, vol.~10, no.~4, pp. 12--25, 2015.

\bibitem{cl4RNN}
A.~Cossu, D.~Bacciu, A.~Carta, V.~Lomonaco, and D.~Bacciu, ``Continual learning
  for recurrent neural networks: an empirical evaluation,'' \emph{CoRR}, vol.
  abs/2103.07492, 2021.

\bibitem{cl4ESN}
A.~Cossu, A.~Carta, C.~Gallicchio, V.~Lomonaco, and D.~Bacciu, ``Continual
  learning with echo state networks,'' in \emph{Proceedings of the 29th
  European Symposium on Artificial Neural Networks, Computational Intelligence
  and Machine Learning (ESANN)}, 2021.

\bibitem{D3.1}
T.~Consortium, ``{Deliverable D3.1: Initial Report on Engineering Methods and
  Architecture Patterns of Dependable CPSoS},'' Tech. Rep., December 2020.

\bibitem{Safecomp21}
G.~Macher, E.~Armengaud, E.~Brenner, and C.~Kreiner, ``{Dependable Integration
  Concepts forHuman-Centric AI-based Systems},'' in \emph{{IN REVIEW SAFECOMP
  2021, Proceedings}}.\hskip 1em plus 0.5em minus 0.4em\relax Springer
  International Publishing AG, 2021.

\bibitem{CHEN2017279}
L.~lan Chen, Y.~Zhao, P.~fei Ye, J.~Zhang, and J.~zhong Zou, ``Detecting
  driving stress in physiological signals based on multimodal feature analysis
  and kernel classifiers,'' \emph{Expert Systems with Applications}, vol.~85,
  pp. 279 -- 291, 2017.

\bibitem{rnn4HSM}
D.~Bacciu, D.~{Di~Sarli}, C.~Gallicchio, A.~Micheli, and N.~Puccinelli,
  ``Benchmarking rc and rnns for human state and activity recognition,''
  \emph{Submitted}, 2021.

\end{thebibliography}
\end{document}